\documentclass{article}
  \usepackage{aaai}
  \title{ Hypothetical revision and matter-of-fact supposition}
  \author{ Horacio Arl\'{o} Costa\thanks{This work was partially supported by a grant from NSF.} \\ Carnegie Mellon University \\ hcosta@andrew.cmu.edu}
  \begin{document}
  \maketitle

\begin{abstract}

The recent literature offers several models of the notion of {\em
matter-of-fact supposition}\footnote{I am using here the terminology
used by James Joyce in \cite{joyce:1999}, pages 182-83.} revealed in
the acceptance of the so-called indicative conditionals.  Some of
those models are qualitative \cite{collins:1990}, \cite{Levi:1996},
\cite{Stalnaker:1984}. Other probabilistic models appeal either to
{\em infinitesimal probability} or {\em two place probability
functions} (\cite{vF:1995}, \cite{joyce:1999}, \cite{mcgee:1994}).
Recent work has made possible to understand which is the exact
qualitative counterpart of the latter probabilistic models
(\cite{vF:1995}, \cite{mcgee:1994},
\cite{Arlo-Thomason:199+}, \cite{Arlo-Parikh:1999}, \cite{Arlo1:1997}. In this
article we show that the qualitative notion of change that thus arises
is {\em hypothetical revision}, a notion previously axiomatized in
\cite{Arlo1:1997} and \cite{Arlo-Thomason:199+}.

This notion is incompatible with AGM as well as with other standard methods of theory change (like Mendelzon and Katsuno's UPDATE).  The way in which matter-of-fact supposition is modeled via hypothetical revision is illustrated via examples.  The model is compared with other qualitative models of indicative supposition (like the one offered in \cite{Levi:1996} and \cite{Stalnaker:1984}), with models of subjunctive supposition, as well as with some of the well know models of learning.  Applications in the theory of games and decisions are considered.
\end{abstract}

\section{Introduction}
The notion of {\em supposition} has a central role in probability theory and the theory of games and decisions.  The notion is usually formalized by appealing to conditional probability.  At each point in time the internal epistemic state of a Bayesian agent is formalized by an unconditional measure encoding its degree of belief and by conditional measures encoding the coherent degree of belief of the agent conditional on suppositions.  The notion of supposition is usually carefully separated form the notion of learning.  Brian Skyrms has stated this distinction as follows in \cite{skyrms:1987}:

\begin{quote}
Updating subjective belief to assimilate a given piece of information and supposing what the world would be like were that bit of information true, are distinct mental processes for which different rules are appropriate...
\end{quote}

Minimizing divergence with respect to a given constraint is the hallmark of supposition, not of learning \cite{joyce:1999}, \cite{diaconis-zabell:1982}. Imposing a constraint is tantamount to fix important features of the final state to which the current state P will be mapped and then minimizing some function measuring divergence from P and any of the possible states exhibiting the selected feature.  

We will not focus here in the probabilistic side of the literature distinguishing learning and supposition.  Reviewing this literature is well beyond the space limitations of this abstract.  Nevertheless the curious reader can consult \cite{joyce:1999}, \cite{Levi:1996},  \cite{Levi:1980}, \cite{skyrms:1987}.  Let me just mention in passing that the distinction between learning and supposing can be obviously made in a purely qualitative setting.  For example, some authors doubt the universality of the so-called axiom of {\em success} establishing that when we revise the current state K with any input A, this input should be part of the revised output K*A.  After all, when an agent is faced with certain evidence, he might decide to reject it.  While this might be true for learning, this is hardly true for supposing.  A moment of reflection will convince the reader that the success axiom is {\em constitutive} of any reasonable notion of supposition.  Something similar happens with the general justification of the operation of contraction.  Why any agent might be persuaded to loose information that at the moment he fully believes?  This has led some researchers to propose that contraction cannot be one of the fundamental operations of theory change.  Again, this is not a problem for the notion of supposition.  An agent can perfectly well engage in a hypothetical exercise that requires contracting his view, while he continues to fully believe what he fully believes.  Considerations of this type can be adduced in order to propose a priori reasonable axioms for supposing.  This will not be, nevertheless, our only way of proceeding here.  We will appeal to work previously done in probability theory (in order to formalize supposition) and then we will determine which are the constraints on qualitative supposition induced by the probabilistic model.  This will require, of course, engaging in the non-trivial task of bridging the notions of probability and belief.  Before plunging into this problem, let me delineate intuitively first which is the type of supposition that we want to capture in this essay. 

It is usually recognized that there are at least two main forms of supposition.  James Joyce puts the contrast between these two salient varieties of supposing in a nice way:

\begin{quote}
Was Hamlet written if Shakespeare did not write it?  Would Hamlet have been written if Shakespeare had not written it?  Unless you are one of those who think that Christopher Marlowe or the earl of Oxford wrote the plays commonly attributed to Shakespeare, you will answer `yes' to the first question and `no' to the second.   Here you are supposing {\em in two distinct ways} that Shakespeare did not write Hamlet.   When you suppose it matter-of-factly you ask how things {\em are} if, as a matter of fact, Shakespeare did not write the play, and, since you know that the play was written by someone, you conclude that it must have been someone other than Shakespeare.  On the other hand, when you suppose that Shakespeare did not write Hamlet in the subjunctive mode you ask how things would have been in  a counterfactual situation in which Shakespeare did not write Hamlet, and, as Shakespeare was a singular genius, you conclude that the play would not have been written at all.  These two species of supposition [...] are represented probabilistically  by Bayesian conditioning and by {\em imaging} [\cite{joyce:1999}, page 182-83].\footnote{The qualitative counterpart of imaging is Mendelzon and Katsuno's UPDATE, a notion well know in the KR community.  Imaging was first proposed in \cite{lewis:1976} as the method of supposition, revealed in the acceptance of counterfactuals.  Again here the idea is to characterize a {\em suppositional} method, rather than an algorithm for learning, following Skyrms' ideas - Lewis, personal communication.}
\end{quote}

The examples used by Joyce were first formulated by Jonathan Bennett and even previously by Ernest Adams in a much quoted essay \cite{adams:1970}.  Adams' example is structurally identical.  Suppose that you are a convinced Warrenite.  Then you will probably accept that `If Oswald didn't kill Kennedy, someone else did'.   Nevertheless this is compatible with your eventual rejection of ` If Oswald hadn't killed Kennedy, someone else would have'.

In this essay we will only deal with the act of supposing revealed by our practices of accepting or rejecting indicative sentences, like ` If Oswald didn't kill Kennedy, someone else did'.  

Two immediate comments are required here.   First we need a qualification of Joyces's point about matter-of-fact supposition.  Joyce's proposal of capturing matter of fact supposition via standard Bayesian conditioning needs a clarification.  Consider the case where a Warrenite accepts `If Oswald didn't kill Kennedy, someone else did'.  This can hardly be captured by saying that the conditional probability of S (someone else aside from Oswald killed Kennedy) given not-O is 1.  In fact, a Warrenite might be convinced that Oswald alone killed Kennedy. Therefore the proposition not-O has, from his point of view zero measure, or infinitesimal measure.\footnote{To be sure there is not need for the Warrenite to assign measure one to `Oswald alone killed Kennedy': O.  But as long as the agent is certain or `almost certain' of O he should do so - see \cite{gardenfors:1988}. There are two alternatives to this view.  A radical one assumes that only logical truths are candidates for full belief. A less radical one (defended in chapter 6 of \cite{maher:1993}) denies that probability one should entail or be entailed by full belief. Neither option is palatable and both have been rejected by the foremost theorists working in the area of belief change.  According to the first option the only interesting and viable notion that epistemologists should study is probability kinematics.  Belief change understood as the change of bodies of full belief containing more than mere tautologies is not a viable enterprise.  The second alternative severs completely the connection between action and belief. Even when an agent declares that he fully believes a proposition, this assertion does not commit him to assign {\em any} degree of belief to the proposition - zero degree is also permissible.}    In any of those cases it seems that we need an extension of conditional probability in order to capture the notion of supposing involved in the acceptance of the conditional in question.  Joyce allows for this extension, as well as other authors (\cite{vF:1995}, for example).  We will appeal to the same idea here.

The second comment is related to a purely qualitative account of these examples.  Some authors propose to formalize matter-of-fact supposition via  AGM  and subjunctive supposition via UPDATE.  Before considering a probabilistic analysis of matter of fact supposition we will argue that an unmodified version of the previous proposal is unfeasible.

\subsection{Matter-of-fact supposition and AGM}

There are several reasons explaining why AGM is inadequate to capture matter-of-fact supposition. In the case of the probabilistic brand of matter-of-fact supposition that we are about to characterize, the conflict between the two notions amounts to logical incompatibility.  Here we will focus on some of the most salient reasons explaining both conflicts.  Others reasons will be provided below.

Let's consider the simple case related to modeling the matter-of-fact supposition involved in accepting `If Oswald didn't kill Kennedy, someone else did'.  The following figure represents a plausible way in which a Warrenite might calibrate his plausibility order, where O is `Oswald alone did shot Kennedy' and S is `Someone else did it'.  The letter J stands for `Lyndon Johnson became president'.

\begin{tabular}{||l|r||}             \hline
$\kappa$       & Possible worlds \\ \hline
0                  & O,  not-S, J \\ \hline
1                  & not-O, S, J \\ \hline
2 & S, O, J \\ \hline
3 & not-S, not-O, not J \\ \hline
\end{tabular}

\centerline{\bf Fig. I.}

A convinced Warrenite can plausibly organize states incompatible with his view as is depicted in Fig. 1.  The immediate rank after the less implausible state (rank 0) is the one in which some person other than Oswald killed Kennedy. Of course there are other possible rankings compatible with the maximal lack of implausibility of (O,  not-S, J) and the acceptance of `If Oswald didn't kill Kennedy, someone else did'.  But in any case (not-O, S) should receive a rank strictly below the one assigned to (not-S, not-O).\footnote{Since the atoms S and O are enough to identify the worlds used in the example, we will  not use J to name them}

So, at first blush it seems that one can rationalize matter of fact supposition via standard AGM methods.  What about the corresponding subjunctive? `If Oswald had not killed Kennedy, someone else would not have done it' can, of course, be  accepted, but acceptance is performed in this second case with respect to a {\em different} ordering.  In this case we need a ranking where (not-S, not-O) is assigned a rank strictly lower than the rank assigned to (not-O, S) - and where the 0-rank coincides with the one in Fig. I.\footnote{Two clarifications. First we follow the standard procedure according to which the rank of a proposition is defined as the lowest rank where the proposition holds.  Secondly, AGM and UPDATE both apply to a general case where the lowest rank encoding beliefs can contain a {\em set} of worlds, rather than just a singleton.  But they can yield different verdicts even when applied to singletons.  This is related to the fact that the intended ontological or causal interpretation of UPDATE requires a notion of similarity that may very well differ from the epistemically motivated metrics required by AGM.}

Perhaps the acceptance of the subjunctive can be rationalized by appealing to an ordering of the type just described.  The previous analysis also shows the interesting fact (sometimes neglected) that matter-of-fact supposition can perfectly well be counter-doxastic.  In other words, although matter-of-fact supposition is a form of supposition that intends to preserve the given facts, in some occasions such type of change requires the formulation of belief-contravening hypothesis.  The problem with matter-of-fact supposition is that only {\em some} counter-doxastic propositions are entertainable, while the supposition of other counter-doxastic propositions leads to incoherence.\footnote{The nature of the beliefs under revision is an issue that will require careful analysis.  Here we will propose to view those changes as expectation-contravening, rather full-belief-contravening.}  This is one of the facts that will show that AGM does not provide the right framework for capturing matter of fact supposition.

Let's consider the former example again.  It is perfectly plausible to suppose subjunctively that nobody killed Kennedy. Moreover in that setting it seems right to accept `If nobody had killed Kennedy, then Lyndon Johnson wouldn't become president'.  But notice that `If nobody did kill Kennedy, then Lyndon Johnson didn't become president' seems absurd (the example is a variant of one recently used by Adams in \cite{adams:1998}, page 281).  Nevertheless if we use the ranking in Fig I.  this conditional should be accepted.  Upon reflection one sees that the failure is radical.  In fact, it seems that one cannot coherently use the matter of fact mood in order to suppose that nobody did kill Kennedy.  This is commonly seen by the fact that people are prone to accept conditionals of the form: `If nobody did kill Kennedy, then I ', where a highly implausible proposition I is used as consequent.\footnote{An example is: `If nobody did kill Kennedy, then I am the Pope'.}

To put it in a more technical manner, matter of fact supposition is not {\em consistency preserving}. The ranking of an agent who engages in matter of fact supposing includes up to rank 2 in Fig. I.  The world in rank 3 is non-entertainable.  Of course `Nobody killed Kennedy' is a consistent proposition.  Nevertheless, when supposed matter-of-factly leads the agent to an incoherent hypothetical state.

The notion of matter-of-fact supposition is rather elusive.  Joyce, for example, tries to characterize it by saying that `the hallmark of matter-of-fact supposition is that it never forces the supposer to revise any of her views about the `facts', that is to say propositions that she is certain are true'.  This prima facie characterization is certainly accurate when one evaluates events of non-zero measure.  But we saw in the previous example that an agent can indeed entertain suppositions (matter-of-factly) carrying zero measure and revise with them the lowest rank of their plausibility measure.  In the example all the measure can be located in rank 0.  Then not-O is a permissible candidate for matter-of-fact supposing, although it carries measure zero. Matter of fact supposition may require to consider how the world {\em is} like when a proposition carrying zero measure is assumed true for the sake of the argument.

The intuition in the example we considered is that there is a set of
`hard facts' about the world that can never be discarded when the view
of the Warrenite is hypothetically changed in a matter-of-fact mood.
The central `hard' fact in the example is the presupposition that {\em
somebody} killed Kennedy -- and Johnson become president.  As Joyce
suggests when one supposes matter-of-factly none of those hard facts
are candidates for revision.  The entertainable propositions are the
ones entailed by the worlds in ranks 0, 1 and 2.  But one can
certainly suppose items compatible with those hard facts.  In doing so
one needs to revise the view encoded in rank 0.  For example, anyone
holding the ranking in Fig. I should accept `If someone else other
than Oswald shot Kennedy then Oswald did not do it'.  This account
presupposes that certainties (full beliefs) are not encoded by the
0-rank.  If this is so which is the attitude encoded in the 0-rank? We
will see that a probabilistic modeling of matter-of-fact supposition
will naturally lead to the idea that the 0-rank encodes an
epistemological notion weaker than certainty, which we will call {\em
expectation}.  The underlying idea is that although expectations and
certainties (full beliefs) carry measure one (or measure infinitely
closed to one) only the latter notion is robust with respect to
suppositions. Only certainties continue to carry measure one in the
presence of {\em any supposition}.

\section{A probabilistic account of matter-of-fact supposition}

Our model uses two-place probability functions, sometimes called Popper functions in the literature.  These functions can be characterized as follows:

A {\em space} is a couple S = $\langle U,F \rangle$ with U a non-empty
set and F a sigma-field on U. We will focus on the conditional measures obeying (I) for any fixed $A$, the function $P(X|A)$ as a function of $X$ is either a (countably additive) probability measure, or has constant value 1 and  (II) $P(B \cap C|A) = P(B|A)P(C|B\cap A)$ for all $A,B,C$ in $F$. The unconditional probability  of $A$, $pr(A)$,  is $P(A|U)$.

Axiom II (usually called the Multiplication Axiom) differs from the usual rule for computing conditional probabilities by allowing to condition on events of measure zero.  Two-place functions have been independently proposed by many authors as the most general probabilistic encoding of matter-of-fact supposition (\cite{joyce:1999} and \cite{vF:1995} are examples).  

In order to determine the qualitative constraints on belief and conditional belief imposed by this account, we need to determine which is the body of belief associated which each two-place measure and how these bodies of belief change when the measures change.  

Paradox-free accounts of the first problem have been considered by van
Fraassen in \cite{vF:1995} and then refined, extended and slightly
modified in \cite{Arlo1:1997} and \cite{Arlo-Parikh:1999}.  The
problem to circumvent here is the so-called lottery paradox
\cite{Kyburg:1961}.  This paradox arises when full belief is
identified with measure one.  We will follow here the proposal in
\cite{Arlo-Parikh:1999}.  According to this account probability one is
a necessary but not a sufficient condition for full belief.

If $P(X|A)$ is a probability measure as a function of $X$, then $A$ is {\em normal} and otherwise {\em abnormal}. In the following we shall confine ourselves to the case where the whole space $U$ is normal and countable.  The notion of normality is closely connected to an epistemic analysis of the notion of {\em a priori}: (A) A is {\em a priori for} P iff P(A $\mid$ X) = 1 for all X, iff U - A is abnormal for P.

A probability {\em core} is a set $K$  which is normal and satisfies the {\em strong superiority condition} (SSC)  i.e. if $A$ is a nonempty subset of $K$ and $B$ is disjoint from $K$, then  $P(B|A \cup B)$ = 0.

The family of cores induced by a two place probability function $P$ is nested.  That this holds in general has been shown in \cite{vF:1995}.  Moreover it can also be shown for spaces of arbitrary size that the chain of belief cores induced by a non-coreless 2-place function P cannot contain an infinitely descending chain of cores (the proof is presented in \cite{Arlo1:1997}).

When the space is countable it can be shown that there is a smallest as well as a largest core.  Moreover, the smallest core has measure 1.  And, more importantly, this innermost core is constituted by all `heavy' points carrying unconditional positive measure (this is showed in \cite{Arlo-Parikh:1999}).  

In \cite{Arlo-Parikh:1999} we proposed that the smallest core be identified with ordinary beliefs (or expectations) and the largest core with full beliefs, so that in general {\em full belief} (certainties/factual constraints) will be distinct from probability 1.  

\begin{flushleft}
{\bf Example I:} The sample space has a countable set of atoms,
resulting from the following: in independent trials, a fair coin is
flipped until we get a head, and then the trials stop.  The set of
possible outcomes is indexed by the number of tails $X = (0, 1, 2,
..., n,... , \omega$), where $\omega$ designates never stopping,
i.e., flipping forever and seeing only tails. Evidently $pr(X$ {\em is
finite}) = 1 and $pr(X = n) = \stackrel{1}{2^{(n+1)}}, (n = 0, 1,
...)$, so that (obviously) $pr(X = \omega$) = 0, and this is the only
null event, apart from the impossible event.
\end{flushleft}

This probability function induces two cores: U and $U - \omega$.  .  It seems unreasonable in this case to require that a rational agent modeled by $P$ ought to {\em fully} believe that $X$ is finite.  But one can say that the agent {\em expects} $X$ to be finite.  

It is easy to build a two-place function having the cores depiected in Fig. I and satisfying: (1) the point (O, not-S) carries measure one (2) the outermost core is 2. Here is a plausible interpretation of this model.  First one can say that the agent in question {\em expects} Oswald to be the sole killer and that, in addition, she fully believes that somebody killed Kennedy.  Her expectations receive measure one.  In this context the proposition that Oswald did not kill Kennedy contradicts the agent's expectations, but it is still compatible with her full beliefs.  According to this view the Warrenite {\em expects} that Oswald is the sole killer, but he is only certain of the fact that somebody killed Kennedy.  The fact that Oswald is the sole killer is still open for revision, but the fact that somebody killed Kenendy is not.  The distinction between expectations and certainties (or factual constraints) is usually made in probability theory by using a different terminology.  The needed distinction is between `almost certain' and certain.  Both notions carry probability one, but only the second is {\em robust with respect to suppositions} in the sense that not only its probability is one but its probability continues to be one in the presence of any supposition.\footnote{van Fraassen characterizes this idea in \cite{vF:1995}  as the probabilistic `a priori'.}

In either case it seems clear that matter-of-fact supposition requires
the specification of two dimensions of qualitative attitudes. One
encoded by the innermost core of P, the other by outermost core.  We
proposed here to interpret the innermost core in such a way that all expectations (almost certainties) are entailed by it.  By the same token certainties (full beliefs/factual constraints) are understood as what is entailed by the outermost core.\footnote{According to this view there are full beliefs but not many of them.  In the example historical facts not only carry
probability one (or probability infinitesimally close to one) but they are also taken as certainties.  Other established facts, like O, still receive maximal measure, but they are only `almost certainties' and are open to matter-of-fact hypothetical revision.}

The system of cores orders the points carrying zero measure outside
the agent's belief set. One can see the points inside the core-system
(receiving positive rank) as carrying infinitesimal probability. The
cores then have the function of ordering those infinitesimals. Of
course it could happen that the outermost core coincides with its
innermost core.  In this case all happens as if no point receives
infinitesimal measure aside the ones in the lowest rank where all
probability mass is concentrated. This is an extreme situation where
`almost certainty' and certainty coincide.  Limit cases of this
sort require further analysis.

Is it possible to explain the eventual acceptance of `If Oswald did not kill Kennedy, someone else did' with respect to a measure whose innermost and outermost cores are rank 0 in Fig. I?  Notice that in this case not only the agent assigns probability one to (O, not-S), but he also refuses to assign infinitesimal measure to any other world. The simple answer in this case is no, the conditional in question should be rejected. Eventual acceptance can only be rationalized in terms of a further change of the established facts. The probabilistic model under consideration cannot consider full-belief-contravening shifts of this sort (the model only covers {\em expansions} of certainties). Models of this type have been considered in \cite{Levi:1996} (see the section on {\em consensus revision}) and to some extent \cite{Stalnaker:1984}.\footnote{See also \cite{vF:1995} where the innermost core of a two-place measure is interpreted as encoding the full beliefs of the agent.}

This essay is only devoted to consider expectation-based models where expectations and full beliefs are distinct. As we will see below, even if this distinction is allowed, a rich model covering sequences of suppositions departs significantly from AGM.

\section{On the dynamics of probabilistic cores}

The problem we need to address now is how the system of cores of a two-place measure changes when the measure is updated. Probabilistically the problem is how to revise the two-place probability P with a proposition A. A natural proposal is the function P(-- $\mid$ -- $\cap$ A), which we can abbreviate by $P^{A}$. Sequences of hypotheses are indicated as follows: $P^{A, B. ...}$.  The following result shows how the system of cores for $P^{A}$ is related to the system of cores for P. 

\vskip 9pt
{\bf  Proposition}
\vskip 9pt

\begin{flushleft}
If $S^{P}$ is the system of cores of P(-- $\mid$ --) and A $\cap$ {\bf U} $\neq$ $\emptyset$, where {\bf U} = $\cup$ $S^{P}$, then all the elements of: S'=  \{C $\cap$ A : C $\in$ S and C $\cap$ A $\neq$ $\emptyset$\}, are {\em exactly} all the elements of the sub-system of belief cores $S^{P^{A}}$ $\cap$ {\bf U}.
\end{flushleft}

\vskip 9pt

Whenever A fails to overlap the largest core of P, or when P is abnormal, we set $P^{A}$ to the abnormal P assigning measure one of every event.  In addition, the innermost core of $P^{A}$, denoted by I(P), is  in this limit case arbitrarily set to $\emptyset$.  If F(P) denotes P's largest core, we also set F(P) to $\emptyset$ by convention.  Then the following properties are satisfied (for non-coreless measures P):

\begin{description}
\item[Expansion] F(P) $\cap$ A = F($P^{A}$)
\item[Success]  I($P^{A}$) entails A
\item[Preservation] If I($P$) $\cap$ A $\neq$ $\emptyset$, then I(P) $\cap$ A = I($P^{A}$)
\item[Restricted Consistency Preservation] If I(P) $\neq$ $\emptyset$, and F(P) $\cap$ A $\neq$ $\emptyset$, then I($P^{A}$) $\neq$ $\emptyset$.
\item[Fixity]  If P is the abnormal measure, I($P^{A}$) = F($P^{A}$) = $\emptyset$.
\item[Cumulativity] I($P^{A, B}$) = I($P^{A \cap B}$)
\end{description}

These properties are the axioms of a notion of change called {\em
hypothetical revision}. The notion has been axiomatized in a more
abstract setting (see \cite{Arlo-Thomason:199+} and
\cite{Arlo1:1997}).  A cumulative model is a tuple $\langle {\bf E},
U, *, \rho,  c \rangle$, where U is a universe of points, {\bf E}
a set of epistemic states, and $\rho$ and {\em c} functions mapping
epistemic states to propositions in U, in such a way that, for each E
in {\bf E}, $\rho$(E) $\subseteq$ {\em c}(E).  Finally * is function
mapping epistemic states and propositions to epistemic states.  We do
not assume that E is closed under *.

The epistemic states in our probabilistic models are instantiated as probability functions, the functions $\rho$ and  {\em c} as our functions I and F, and * is the analogue of the [...] operation on probability functions.  Then the property $\rho$(E) $\subseteq$ {\em c}(E) is automatically satisfied.  The rest of the properties of hypothetical revision are also satisfied via a straightforward translation.  For example, Expansion gets translated into {\em c}(E) $\cap$  A = {\em c}(E * A).

\vskip 9pt

\begin{flushleft}
[Expansion] {\em c}(E) $\cap$ A = {\em c}(E*A)
\end{flushleft}

\begin{flushleft}
{\em Explanation}:  Matter of fact supposition is characterized by the fact that certainties (full beliefs/factual constraints) are always preserved.  Suppositions  contravening the factual constraints encoded in {\em c}(E) lead to incoherence. Suppositions compatible with established facts in {\em c}(E) go by the mere expansion of {\em c}(E) with the supposed item.
\end{flushleft}

\vskip 9pt

\begin{flushleft}
[Success] $\rho$(E*A) entails A
\end{flushleft}

\begin{flushleft}
{\em Explanation}:  This axiom seems constitutive of any notion of supposing.  To be sure not all forms of belief change need to be successful.  Incoming information may or may not to be incorporated in the current view.  But supposing A requires to consider a hypothetical state where A holds for the sake of the argument.
\end{flushleft}

\vskip 9pt

\begin{flushleft}
[Preservation] If $\rho$(E)  $\cap$ A $\neq$ $\emptyset$, then $\rho$(E) $\cap$ A = $\rho$(E*A)
\end{flushleft}

\begin{flushleft}
{\em Explanation}:  This requirement reflects Ramsey's ideas about acceptance of conditionals.  When the agent is open about a hypothetical input the supposition goes by simple conditioning.
\end{flushleft}

\vskip 9pt

\begin{flushleft}
[Restricted Consistency Preservation] If $\rho$(E)  $\neq$ $\emptyset$, and {\em c}(E) $\cap$ A $\neq$ $\emptyset$, then $\rho$(E*A) $\neq$ $\emptyset$.
\end{flushleft}

\begin{flushleft}
{\em Explanation}: Suppositions with items compatible with the set of factual constraints {\em c}(E) lead to coherent suppositional scenarios.
\end{flushleft}

\vskip 9pt

\begin{flushleft}
[Fixity] If $\rho$(E) = $\emptyset$ = {\em c}(E), then $\rho$(E*A) = $\emptyset$. 
\end{flushleft}

\begin{flushleft}
{\em Explanation}: Supposition does not have the goal of restoring consistency to an eventually incoherent view.  The act of supposing with respect to an incoherent view is vacuous.
\end{flushleft}

\vskip 9pt

\begin{flushleft}
[Cumulativity] $\rho$((E*A)*B) = $\rho(E*(A \cap B))$
\end{flushleft}

\begin{flushleft}
{\em Explanation}: This axiom reflects the idea that the conditions of acceptance of  `If A, then if B, then C' are identical  with the ones induced by `If A and B, then C'.\footnote{Notice that this idea {\em does not} reflect the view that `If A, then if B, then C' is the deceptive linguistic expression of `If A and B, then C'.  This view is compatible with the popular view among probabilists that there are not iterated conditionals.  The view encode in cumulativity is that there are bona fide iterated conditionals  but that their conditions of acceptance are regimented in a particualr manner.}
\end{flushleft}

\subsection{Darwiche and Pearl's axioms and Hypothetical Revision}

Let a cumulative model be called {\em universal}, when for every epistemic state E, {\em c}(E) = U.  Let a cumulative model be called {\em consistent} if for each E, $\rho$(E) is non-empty.  It is easy to see that the following properties hold in all universal and consistent models. 

\begin{description}
\item[Success]  $\rho(E*A) \subseteq A$
\item[Preservation] If $\rho$(E)  $\cap$ A $\neq$ $\emptyset$, then $\rho$(E) $\cap$ A = $\rho$(E*A)
\item[Consistency Preservation] $\rho(E*A)$ is consistent if $A$ is.
\item[Conjunctive Revision] If $\rho(E*A)\cap B \neq \emptyset$, then $\rho(E*A)\cap B = \rho(E*(A \cap B))$
\end{description}

Conjunctive revision follows from Cumulativity and Preservation.  The previous properties are AGM-like properties in the iterated setting studied by Darwiche and Pearl \cite{darwiche-pearl:1997}.  In other words, the non-iterated properties of universal and consistent models of hypothetical revision exhibit AGM-like properties.\footnote{We are adopting here, nevertheless, a Darwiche-Pearl type of axiomatization, where the central notion is the notion of {\em epistemic state} rather than the notion of {\em belief set} - see the discussion in \cite{darwiche-pearl:1997}}.  Of course, when universality (or consistency) is lifted, we have a first discrepancy with the Darwiche-Pearl axioms, because Consistency Preservation holds in this case only in its restricted version. 
 
The cumulative models characterize the generalized Horn patterns of inference of the {\em Rational} system of non-monotonic consequence.  This is shown in \cite{Arlo-Parikh:1999}.  This should perhaps be expected taking into account the relationships between AGM-like postulates and rational inference, and the fact that this notion of inference has no counterpart for consistency preservation.  

The fact that universal and consistent models obey AGM-like properties, and that cumulative properties validate well known non-monotonic axioms should not obscure the point that, when iteration is allowed, hypothetical revision proceeds in a manner clearly at odds with most methods of {\em belief change} studied in the literature.  The central issue is the property of Global Success, mentioned above. In fact, one of the crucial features of the probabilistic notion of supposition under consideration is that when a suppositional constraint is adopted it is then {\em held fixed} during subsequent changes.  

\begin{description}
\item[Global Success]  $\rho((E*A)*B) \subseteq A$\footnote{ John Collins in \cite{collins:1990} has argued in favor of using AGM methods in order to represent matter-of-fact supposition.  Nevertheless, in section 3.8 of \cite{collins:1990} Collins restricts the set of permissible entrenchments in order to accommodate Global Success.  Although he does not offer an axiomatization, it seems that the resulting model is an instance of hypothetical revision.}
\end{description}

\subsection{Hajek and Harper on probabilistic supposition}

\cite{hajek-harper:1996} contains one of the few qualitative set of (non-iterated) axioms for the notion of qualitative supposition induced by two-place functions.  In \cite{Arlo-Thomason:199+} it is shown that a modified version of these axioms follow as theorems from our more general axiomatization:

\begin{description}
\item[(E1)] $\rho(E*A)\subseteq A$.
\item[(E2)] If $\rho(E)\subseteq A$, then $\rho(E)=\rho(E*A)$.  
\item[(E3)] If $A\subseteq B$ and $\rho(E*B)\cap A\neq\emptyset$,
	then $\rho(E*A)=\rho(E*B)\cap A$. 
\item[(E4] If $A\subseteq B$ and $\rho(E*B)=\emptyset$,
	 then $\rho(E*A)=\emptyset$.
 \end{description}

\section{Non-Archimedean subjective probability and two-place probability functions}

The following sections will review some of the important applications of the account presented above. But before considering this issue it would be appropriate to outline the connections between `infinitesimal' probability and the type of generalized conditional probability just considered.

There are two standard manners of representing conditioning with zero
measure events. One strategy has already been reviewed and used to
model matter-of-fact supposition (two-place probability functions).
An alternative method has gained currency during the last half of the
past century. The gist of this alternative move is to propose that a
proposition can carry {\em infinitesimal} value greater than zero, but
smaller than any standard number. Instead of assigning zero to the
proposition that a randomly chosen point is in the Western hemisphere,
we give to it an infinitesimal value.  So, if we assign an
infinitesimal value {\em i} to the point being in the Southern part of
the Western hemisphere and 2{\em i} to the proposition that it is in
the Western hemisphere; the proposition that it is in the Southern
part of the Western hemisphere, conditional on its being in the
Western hemisphere, is defined and carries infinitesimal value {\em
i}/2{\em i}.  Of course, in order to have a workable proposal, a
calculus of infinitesimal magnitudes has to be developed.

Several well know properties of ordered algebraic fields are violated.
The so-called {\em Archimedean} axiom is violated.  If we focus on the
real field, the axiom establishes that, for any positive number {\em
r}, no matter how small, there is an integer {\em n} such that {\em
n}{\em r} $>$ 1. A non-Archimedean field will contain at least an
`infinitesimal' element $\varepsilon$ with the property that {\em
n}$\varepsilon$ $<$ 1, for every positive integer {\em n}, even though
$\varepsilon$ is itself positive. Such a number should be smaller than
any real and it cannot be a real itself.  Abraham Robinson's pioneer
work in this area made possible the articulation of a mathematically
workable theory of non-Archimedean real fields.  Since then the theory
has been variously applied in the theory of games and decisions.
Non-Archimedean expected utility and subjective probability have been
studied and variously applied.  More recently this account has been
considered in KR applications \cite{lehmann:1998}.

The two approaches just considered (infinitesimal probability and two-place functions) have grown independently and their foundational problems have different roots. Perhaps non-standard probability is a more adequate (or `intuitive') representational tool, but two-place functions have the additional advantage of a simple axiomatization - at last this is so for most applications where the domain of the underlying sigma field is at most countable. In addition, as we showed above, bridges between qualitative and quantitative change can be established for two-place functions.

In spite of the aforementioned differences Van McGee has recently offered an interesting result showing that it is possible to map statements about non-standard probabilities into corresponding facts about ordinary valued two-place probability functions \cite{mcgee:1994}.  The gist of the result is a back and forth proof showing that the standard values of infinitesimal probability functions are representable as two-place functions, and that every two-place function is representable in terms of the standard real values of some infinitesimal measure.  When this theorem is complemented by the connection between qualitative belief change and (two-place) probability kinematics presented above it gives an interesting insight about the qualitative commitments of models of supposition in terms of infinitesimal probability.  

\subsection{Applications: non-monotonic reasoning}

Judea Pearl and Lehmann-Magidor provided the first probabilistic
models of non-monotonic reasoning in terms of infinitesimal
probability.  Pearl also offered the first accounts of several
`qualitative' methods of belief revision (like Spohn's {\em ordinal
conditional funcitons}) as offering an infinitesimal analysis of
conditional probabilities \cite{pearl:1994}.  Taking into account
McGee's result, the analysis presented in this paper should prompt the
conjecture that some of the standard systems of non-monotonic logic
should admit models in terms of matter-of-fact supposing (via
two-place functions).  The conjecture has been studied and confirmed
in \cite{Arlo-Parikh:1999}.  Unlike previous models this work produces
an analysis of non-monotonic consequence in terms of probabilistic
{\em expectations} - unifying existing probabilistic models with
purely qualitative models in terms of expectations
\cite{gardenfors:1994}.  The system of {\em rational logic} of
Lehmann-Magidor is characterized in terms of the universal and
consistent probabilistic models introduced above.

The gist of the analysis is as follows.  B follows non-monotonically from A with respect to the two-place measure P(X $\mid$ Y) whenever P(B $\mid$ A) = 1.  Moreover P(B $\mid$ A) = 1 if and only of the smallest core of P(X $\mid$ Y $\cap$ A) entails B.  In other words, B follows non-monotonically from A with respect to the two-place measure P(X $\mid$ Y) if and only if the expectations associated with P(X $\mid$ Y $\cap$ A) entail B.

\subsection{Applications:Iterated probabilistic conditionals} 

McGee's result is presented only for non-iterated probability change. In \cite{Arlo-Thomason:199+} it is shown that (1) McGee's canonical proof cannot be extended in order to study sequences of changes, (2) that there is nevertheless a manner of extending the proof, and (3) that this extension requires the assumption of Cumulativity.  

This result is suggestive. It shows that Cumulativity is a qualitative property deeply rooted in the two types of non-standard conditional probability considered above, and that it is needed in order to guarantee their coincidence in the iterated case. This is so quite independently of the reasons pro and con one might adduce in order to use it in models of supposition.  
 
Since Cumulativity has been rejected in most of the contemporary models of inquiry (AGM and UPDATE are examples) the former results suggests that a unified picture of infinitesimal and two-place probability might be only compatible with certain interpretations of those extensions of conditional probability. Matter-of-fact supposition is, of course, one of the possible intended interpretations. 

\section{More applications: Decision Theory}

The distinction between matter-of-fact supposition and subjunctive supposition is crucial to understand the current debate about the foundations of decision theory.  Most of the so-called {\em causal} decision theorists defend the idea that the notion of conditional probability used in the calculation of expected utility goes by imaging (or UPDATE).  In other words they argue that decision makers considering hypothetical situations engage in an exercise of subjunctive, rather than matter-of-fact supposition (\cite{joyce:1999} offers an up to date overview of work in this field together with novel results).  Judea Pearl is the most prominent KR researcher working in this field.  He has recently written a book-length argument in defense of the causal theory \cite{pearl:2000}. He has also produced arguments suggesting that subjunctive supposition can be understood in terms of matter-of-fact supposing in the context of a richer theory where causal influence is encoded via graphs \cite{g-pearl:1992}. Meek and  Glymour have offered a slightly different reduction.  The basic idea is that supposing subjunctively that A is tantamount to suppose matter-of-factly that an action has been performed in order to bring A about \cite{Meek:1994}. Finally Levi considered a reduction of different sort in \cite{Levi:1996}.

The second application in this field is related to models of non-Archimedean subjective probability, particularly conditional non-Archimedean probability.  The results presented above suggest that this notion can be understood in terms of matter-of-fact supposition.  

\section{More applications: Game Theory}

The probabilistic encoding of the notion of matter-of-fact supposition
plays a central role in recent extensions of Harsanyi's theory of
types in game theory. For example, Battigalli and Sinischalchi
proposed in \cite{Battigalli:1998} to use two-place functions in order to
define hierarchies of conditional types in multistage games. Dov Samet
has simultaneously studied qualitative notions of hypothetical
knowledge in games.  These applications require the use of iteration.
Nevertheless it is often not sufficiently appreciated that in these
settings the qualitative properties of probabilistic models in terms
of two-place functions are at odds with most of the standard models of
belief change. In this essay we offered an exact modeling of those
qualitative properties, some intuitive support for them and a
comparison with other well-known methods.

\section{Acknowledgements}

Thanks are due to Isaac Levi, Rohit Parikh, Judea Pearl, Teddy Seidenfeld, Richmond Thomason and Bas van Fraassen. I would like to thank as well two anonymous referees for their helpful comments. 

\bibliographystyle{aaai}
\bibliography{br}

\end{document}